\documentclass[conference]{IEEEtran}
\IEEEoverridecommandlockouts
\usepackage{cite}
\usepackage{amsmath,amssymb,amsfonts}
\usepackage{algorithmic}
\usepackage{graphicx}
\usepackage{textcomp}
\usepackage{xcolor}
\usepackage[hidelinks]{hyperref}
\usepackage{booktabs}
\usepackage{comment}
\usepackage{float}

\renewenvironment{quote}{%
  \list{}{%
    \leftmargin0.25cm   
    \rightmargin\leftmargin
  }
  \item\relax
}
{\endlist}

\def\BibTeX{{\rm B\kern-.05em{\sc i\kern-.025em b}\kern-.08em
    T\kern-.1667em\lower.7ex\hbox{E}\kern-.125emX}}
\begin{document}

\title{What Can You Say to a Robot? Capability Communication Leads to More Natural Conversations
\thanks{This research was (partially) funded by the Hybrid Intelligence Center, a 10-year programme funded by the Dutch Ministry of Education, Culture and Science through the Netherlands Organisation for Scientific Research, \url{https://hybrid-intelligence-centre.nl}, grant number 024.004.022.}
}

\author{\IEEEauthorblockN{Merle M. Reimann}
\IEEEauthorblockA{
\textit{Vrije Universiteit Amsterdam}\\
The Netherlands \\
\href{https://orcid.org/0000-0003-3076-5402}{0000-0003-3076-5402}} 
\and
\IEEEauthorblockN{Koen V. Hindriks}
\IEEEauthorblockA{
\textit{Vrije Universiteit Amsterdam}\\
The Netherlands \\
\href{https://orcid.org/0000-0002-5707-5236}{0000-0002-5707-5236}}
\and
\IEEEauthorblockN{Florian A. Kunneman}
\IEEEauthorblockA{
\textit{Utrecht University}\\
The Netherlands \\
\href{https://orcid.org/0000-0002-1932-3200}{0000-0002-1932-3200}}
\and
\IEEEauthorblockN{Catharine Oertel}
\IEEEauthorblockA{
\textit{Delft University of Technology}\\
The Netherlands \\
\href{https://orcid.org/0000-0002-8273-0132}{0000-0002-8273-0132}}
\and
\IEEEauthorblockN{Gabriel Skantze}
\IEEEauthorblockA{
\textit{KTH Royal Institute of Technology}\\
Sweden \\
\href{https://orcid.org/0000-0002-8579-1790}{0000-0002-8579-1790}}
\and
\IEEEauthorblockN{Iolanda Leite}
\IEEEauthorblockA{
\textit{KTH Royal Institute of Technology}\\
Sweden \\
\href{https://orcid.org/0000-0002-2212-4325}{0000-0002-2212-4325}}
}

\maketitle

\begin{abstract}
When encountering a robot in the wild, it is not inherently clear to human users what the robot's capabilities are. When encountering misunderstandings or problems in spoken interaction, robots often just apologize and move on, without additional effort to make sure the user understands what happened. We set out to compare the effect of two speech based capability communication strategies (proactive, reactive) to a robot without such a strategy, in regard to the user's rating of and their behavior during the interaction. For this, we conducted an in-person user study with 120 participants who had three speech-based interactions with a social robot in a restaurant setting. Our results suggest that users preferred the robot communicating its capabilities proactively and adjusted their behavior in those interactions, using a more conversational interaction style while also enjoying the interaction more.

\end{abstract}

\begin{IEEEkeywords}
Human-robot-interaction; dialogue management; spoken interaction; user study
\end{IEEEkeywords}

\section{Introduction}
Although social robots have seen increasing use in recent years, for example in airports \cite{tonkin2018design}, supermarkets \cite{iwamura2011elderly}, and schools \cite{benitti2012exploring}, they are still not common everyday interfaces for most people. Because of this, people do not know how they can best interact with a robot they meet for the first time. When acquiring new devices, they usually come with a manual; however, this is not practical for devices encountered in the wild. Due to the wide variety of robot morphologies and interaction modalities, it is difficult to generalize experiences made during one interaction to a different robot or interaction. 

In human-robot interaction (HRI), speech offers a flexible and hands-free method of interaction. Even for speech-based robots, the type of dialogue manager and the robots' capabilities vary \cite{reimann2024survey}. This variation makes it difficult for users to know what to expect while interacting with a robot. Even if a robot with the same embodiment is used, it is not likely that it has the same, or even similar, capabilities. This can lead to incorrect expectations if a person assumes that they know how to interact with this type of robot. To help users better understand the robot and its actions during an interaction, increased transparency is one of the methods used.
To increase the transparency of a system the robot can be made more explainable \cite{anjomshoae2019explainable}. Other methods of managing the user's expectations include choosing the embodiment and voice in a way that fits with the actual capabilities \cite{marge2022spoken} or by discussing the expectations directly with the users \cite{ligthart2017expectation}.

In this paper, we look at how a robot can communicate its capabilities through the interaction itself and by that increase the transparency regarding its actual capabilities.
For this, we conducted and analyzed an in-person user study with 120 participants in a restaurant setting, where people were asked to get seated and order food by interacting with a robot waiter. The participants interacted with a robot using one of two capability communication strategies or a baseline one. One proactively communicates the capabilities of the robot, while the other one focuses on communicating them when the robot detects problems. We contribute to the improvement of transparency in HRI by designing two different capability communication strategies and showing how they influence the interaction and the user's perception of it.

\section{Related work}

\subsection{Spoken human-robot interaction}

When interacting with robots, their embodiment and the situated nature of the interaction add additional requirements to the dialogue system \cite{jokinen2018dialogue}. Various approaches for dialogue management have been used for human-robot spoken interaction in recent years, ranging from hand-crafted to data-driven approaches, or a hybrid combination of both \cite{reimann2024survey}. Although many of the robots used for spoken HRI are humanoids, other morphologies are also used for speech-based interactions \cite{reimann2024survey}. 
The morphology of the robot plays a role in how people perceive and evaluate it \cite{kunold2023not}.

For spoken human-robot interaction, speech recognition is still a challenge, especially if the robot is deployed in noisy environments or is making noise by moving. Strategies that are used to mitigate that effect, like the use of wakewords, are influencing the user behavior \cite{wen2023fresh}.
Moore \cite{moore2015talking} argues that just hearing and producing speech is not enough to create intelligent communicative robots, but that understanding is crucial.
As robots become more capable and are used in more complex scenarios, the requirements for spoken interaction increase as well \cite{marge2022spoken}. Especially with Large Language Models (LLMs) becoming more common, this can influence the user's expectations of the robot's capabilities.

\subsection{Transparency in human-robot interaction}

Tulli and colleagues \cite{Tulli_Correia_Mascarenhas_Gomes_Melo_Paiva_2019} define transparency as ``an appropriate mutual understanding and trust that leads to effective collaboration between humans and agents.'' 
Transparent systems often focus on making the robot's behavior understandable to the user \cite{Schött_Amin_Butz_2023} and are used in various domains, such as teaching robots \cite{Hirschmanner_Gross_Zafari_Krenn_Neubarth_Vincze_2021} and healthcare \cite{Nesset_Robb_Lopes_Hastie_2021}. Explainability is used to make systems more understandable and communicate their capabilities, by providing explanations of the shown behavior. Capability communication is a form of self-explanation, which aims at enhancing the transparency of a system.
The types of explanations people prefer vary between situations, with why-explanations being highly rated for most scenarios \cite{Wachowiak_Fenn_Kamran_Coles_Celiktutan_Canal_2024}.
Kulesza et al. \cite{Kulesza_Stumpf_Burnett_Yang_Kwan_Wong_2013} found that the completeness of the explanations helps users form more accurate mental models.
The use of different (un)solicited pre-/post-corrective communication strategies does impact the level of trust users have if failures happen \cite{ye2019human}.

Although the efficiency of the interactions depends on the transparency \cite{lyons2013being}, it is still unclear which level of transparency is the best \cite{Bhaskara_Skinner_Loft_2020}, and which one is needed to not hinder the interaction through too much transparency \cite{fischer2018transparent}.
For the evaluation of transparent systems, trust, and performance are commonly used metrics \cite{Schött_Amin_Butz_2023,hoffman2023measures}. For the perceived usability of transparent systems, the literature shows inconsistent conclusions, with some studies indicating an increase in usability, while others do not \cite{Bhaskara_Skinner_Loft_2020}.

Explanation generation is especially important in the context of failures, where it can be used to explain the failure to the user \cite{diehl2022did, das2021explainable, khanna2023user, lemasurier2024reactive}. Those failure explanations should be understandable also by non-experts who are using the robot \cite{das2021explainable}. The explanations can include various levels of detail, by including a possible failure resolution, the cause of the failure, and examples of previous successful actions \cite{khanna2023user}. Different strategies have been used to deal with failures also in non-conversational ways \cite{reig2021flailing}. In contrast to the work on failures, we are not looking at failure resolution strategies, but at communication of capabilities in general.

\section{The robots' capability communication strategies}
We propose two different capability communication strategies. Those strategies provide additional information to the user at different points during the interaction with the aim of increasing the transparency of the system's capabilities and helping users understand how they can use the robot.
In our study, the additional information is either provided proactively or reactively. While previous research has explored proactive and reactive behaviors, our study is the first as far as we know, to directly compare these specific strategies for general capability communication in terms of their effect on user interaction styles and preferences.

\paragraph{Baseline} The first strategy (``baseline'') had basic repair strategies in cases of misunderstandings, without a focus on capability communication. If something goes wrong in the interaction and the robot is able to detect it, it apologizes for the problem and then continues with the dialogue, like it is often done by commercial systems.

\paragraph{Reactive} The second strategy (``reactive'') uses repair strategies to communicate its conversational capabilities to the user. For this, we use intent classification and the confidence values of the speech recognition system. The identified intents and their confidence values, together with the conversation history and instructions, are used to prompt an LLM (GPT4) to generate a response. If the robot detects a repair request, it focuses on resolving it before continuing with the conversation. In cases where the confidence value of the speech recognition is below a threshold, a repair sequence is triggered by the robot itself. The robot-initiated repair is also used for utterances not classified as one of the intents. In both cases, the robot tells the user what it understood and what it was expecting in this situation.

\paragraph{Proactive} The third strategy (``proactive'') focuses on communicating the capabilities in advance. The robot mentions the capabilities relevant for the interaction, before starting the interaction part. The authors of \cite{van2024proactive} define proactive HRI as ``HRI where the
robot exhibits some level of either anticipatory or situation controlling capabilities". In our work we focus on the situation controlling capabilities of the system.

\subsection{Dialogue examples}
\label{sec:examples}
The following transcript examples show the differences between the conditions, as described above. The examples are just created for illustrative purposes:

\paragraph{Baseline condition} 
\begin{quote}
\textbf{Robot:} Welcome to our restaurant. Can you please tell me how many people are in your party?

\textbf{User:} Mirchi

\textbf{Robot:} I'm sorry, I didn't catch that. Could you please tell me how many people are in your party?
\end{quote}

\paragraph{Reactive condition}
\begin{quote}
\textbf{Robot:} Welcome to our restaurant. Can you please tell me how many people are in your party?

\textbf{User:} Mirchi

\textbf{Robot:} I'm sorry, my speech recognition failed and I understood "Mirchi". Could you tell me again how many people you are in total?
\end{quote}

\paragraph{Procative condition}
\begin{quote}
\textbf{Robot:} Welcome to the restaurant. I will seat you and take your order. Please understand that I am a robot and, therefore, might experience some problems understanding your utterances. In those cases, please give me the information again. I can help with small questions, but I work best, if we stick to the conversation flow. Let me start: How many people are you in total?

\textbf{User:} Mirchi

\textbf{Robot:} I'm sorry, I didn't catch that. Could you please tell me how many people are in your party?
\end{quote}

\section{Hypotheses}

Since transparency is increasing the efficiency of interactions \cite{lyons2013being}, we assume that for our task-based interaction, users prefer a robot that provides more information about its capabilities over a robot that does not provide any additional information.

\noindent
\textbf{H1}: The robot that communicates its capabilities reactively is preferred over the robot that is not communicating its capabilities.

\noindent
\textbf{H2}: The robot that communicates its capabilities proactively is preferred over the robot that is not communicating its capabilities.\\

\vspace{-1em}
The form of communication influences the perception of an interaction and perception of the robot \cite{ye2019human}.
LeMasurier and colleagues \cite{lemasurier2024reactive} found a preference for proactive over reactive explanations for failures, with the proactive system being perceived as more trustworthy and intelligent, than the reactive one. Based on this, we assume that there is also a difference in preference for capability communication.

\noindent
\textbf{H3}: There is a difference between the reactive and the proactive capability communication strategy.\\

\vspace{-1em}
For all the hypotheses specified above, we look at the differences in regard to
\begin{enumerate}
    \item the user satisfaction,
    \item the ease of use,
    \item the user’s trust in the performance of the robot, and
    \item the user's willingness to use the robot again.
\end{enumerate}

In addition to subjective measures, we also explore quantitative measures (see Sec.~\ref{sec:measures}) to further understand the possible differences between the conditions.

\section{Methods}
To investigate whether the robot's communication about its capabilities influences the user's perception of the robot and its abilities, we ran a user study. We compared two capability communication strategies with a baseline in a restaurant setting. The participants acted as guests in a fictional restaurant, where the robot served as a service robot for seating and taking orders.
The study was carried out following the ethical guidelines of KTH, where the study was conducted. An additional ethics self-check was performed at the Vrije Universiteit Amsterdam. The study was pre-registered\footnote{\url{https://osf.io/qdm8a/?view_only=ece09ff3713c417490ade395c19da483}}.

\subsection{Participants}
An a priori power analysis using G*Power \cite{faul2007g} for fixed effect, omnibus, one-way ANOVA indicated that, based on a moderate effect size, a minimum sample of 111 participants was required. We recruited a total of 120 participants aged 18 to 63 years ($M = 29.8, SD = 9.11$). Of the 120 participants, 49 identified as female, 70 as male, and 1 preferred not to disclose their gender. The participants had various degrees of experience with social robots ($M = 2.48, SD = 1.11$ on a self-report 5-point Likert scale, baseline = [11,7,13,8,1], reactive = [13,10,9,3,1], proactive = [3,15,12,4,3]).

Participants were recruited through mailing lists, Slack workspaces, flyers, 
a website for participant recruitment, direct recruitment on campus, a web portal for master's students, and word of mouth. All participants received a 100 SEK gift card for their participation. We recruited participants without severe hearing or speech impairments, who were fluent in English, since the study required speech-based interaction.

\subsection{Design}
\label{sec:design}
\begin{figure}[t]
    \centering
    \vspace{-2em}
    \includegraphics[width=0.85\linewidth]{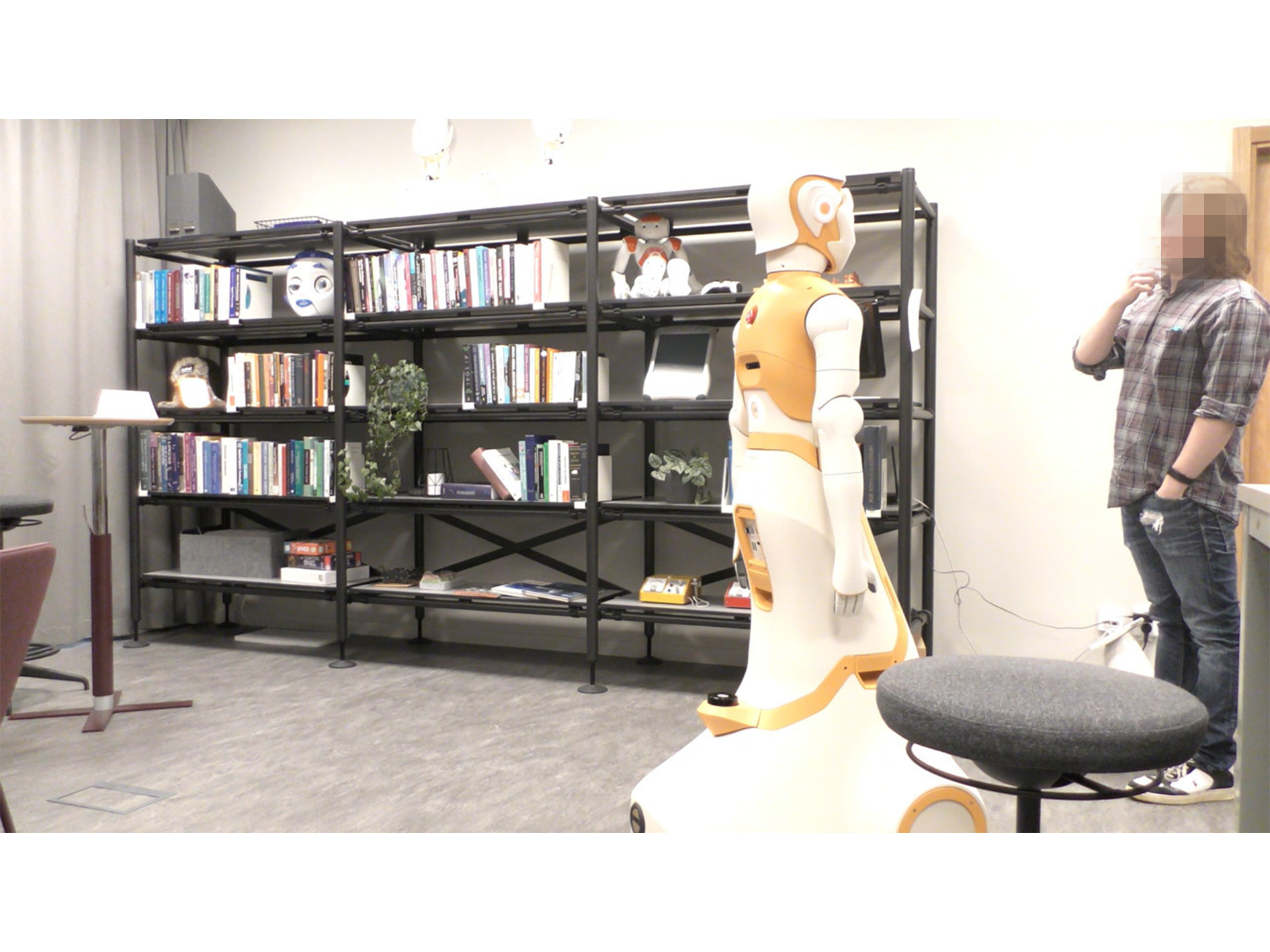}
    \vspace{-2em}
    \caption{The setup of the experiment. ARI greets the participant and then leads them to the high table at the left.}
    \vspace{-1.5em}
    \label{fig:setup}
\end{figure}

The study had three conditions and a between groups design. The conditions differed in the capability communication strategy, which was our independent variable. All participants completed a total of three interactions with different tasks. For the first interaction, there was no guidance for the participants apart from being asked to act like they would in a restaurant. This was done to see how people interact with the robot if they can choose their actions freely. In the following two interactions, we gave additional instructions to participants to create opportunities for a variety of language-based interactions (see Tab.~\ref{tab:scenarios}). The scenarios were chosen in such a way that they provided the participants with a task that they could solve in different ways. There was not just one solution for each scenario; they were intended to help users test the capabilities and limitations of the system. The robot's interaction strategy remained the same throughout all three interactions with each participant, depending on the conditions participants were assigned to. As a dependent variable, we recorded user enjoyment, perceived ease of use, performance trust, willingness to use the robot again, and the understanding of the robot's behavior using questionnaires, as well as objective data from logs.

\subsection{Materials and measures}
\label{sec:measures}
We used an ARI robot from PAL robotics for the experiment. ARI is a 1.6m tall mobile humanoid robot (see Fig.~\ref{fig:setup}), which interacted fully autonomously in our experiment. Google Dialogflow was used for speech recognition and intent classification. Each user utterance intent was classified as \textit{getSeated}, \textit{order}, \textit{repair}, or \textit{DefaultFallback}.
The text response produced by prompting an LLM (\href{https://openai.com/index/gpt-4/}{GPT-4}) was then converted to speech using ARI's text-to-speech conversion. Each condition used one LLM prompt for seating and another prompt for ordering, leading to a total of 6 prompts. The prompts were refined during a pilot study with 8 participants and can be found in the supplementary material.


\paragraph{Questionnaire}
For the demographics, we recorded the age, gender, and previous experience with social robots. 
While we focus on communicating the capabilities, we do not want to negatively impact the overall quality of the interaction with our approaches. Therefore, we measured

\begin{enumerate}
    \item enjoyment,
    \item ease of use,
    \item willingness to use the robot again,
    \item performance trust and
    \item the perceived understanding of the robot’s behavior.
\end{enumerate}

We used standardized questionnaires to measure user enjoyment (5-point Likert scale, $\alpha>0.83$) and perceived ease of use (5-point Likert scale, $\alpha>0.83$) \cite{heerink2010assessing} as well as performance trust (7-point Likert scale, $\alpha>0.84$) \cite{malle2021multidimensional}. All questionnaires showed good internal consistency, with Cronbach's Alpha greater than 0.8. The willingness to use the robot again (``I would use the robot again'') and the perceived understanding of the robot's behavior (``I feel like I understand the robot's behavior'') were single items.

Trust is commonly used to evaluate transparent systems \cite{Schött_Amin_Butz_2023}. Since there is no consensus on whether transparency helps with perceived usability \cite{Bhaskara_Skinner_Loft_2020}, we included a questionnaire on ease of use. 
In addition to this, we also asked the participants to rate their experience with social robots, to see whether the effect of the strategies differs for people with less/more expertise.

\paragraph{Dialogue}
In addition to the questionnaire responses, we recorded interaction logs containing the transcripts recorded from the user responses and the robot utterances.
Since all conditions have the same topic and requirements, we can see whether the conditions influence the users' behaviors and choice of words. For this we looked at (1) the number of words per utterance, (2) the number of unique words, (3) the number of utterances per condition and (4) the dialogue acts.

\paragraph{Videos}
We recorded a video of the interactions, to be able to check the original utterances.
To gain insights into the problems and their distribution across the conditions, we analyzed a random subset of the videos. In our analysis we focused on (1) misunderstandings and their severity, (2) non-understanding, (3) missed beginnings of utterances and their severity, (4) task success, (5) repair and (6) other observations made.

\subsection{Procedure}
\label{sec:procedures}
We ran the experiment in a smart kitchen lab, to reduce the feeling of the participants of being in a lab.
After reading the information sheet about the study, participants were asked to fill in a consent form. Thereafter, they were presented with a general introduction to the task and its setting. After the first interaction, participants were asked to complete the first questionnaire. For the second and third interactions, we provided the participants with a scenario (see Tab.~\ref{tab:scenarios}) and asked them to interact with the robot according to it.
After completing all interactions, participants filled in the questionnaire again. Since we used a between-groups design, all three interactions were in the same condition.

\begin{table}[t]
    \centering
    \caption{The two scenarios the users were asked to keep in mind during the second and third interaction.}
    \label{tab:scenarios}
    \begin{tabular}{p{0.4\textwidth}}
    \toprule
        Scenario 1\\
        \midrule 
         After a long and stressful day, you are planning to treat yourself to a nice meal. Since you have been super busy all day long, you did not have any lunch and are quite hungry now. You are going to the restaurant in the hope to eat their best dish(es).\\
         \\
         \toprule
         Scenario 2 \\
         \midrule
         You are planning to visit the restaurant together with two friends. However, since they are running late, they asked you to already go ahead and order for all of you. One of them is a vegan, while the other friend is gluten intolerant.\\
         \bottomrule
    \end{tabular}
    \vspace{-2em}
\end{table}

During the interactions with the robot, the participant was alone in the room, but a researcher was constantly monitoring the interaction from an observation room next door and could intervene if necessary, which was not the case for the participants included in the analysis.

\section{Results}

We excluded seven participants. Six due to technical problems and one due to an incomplete second questionnaire, leaving us with 113 participants for the analysis.

\subsection{Quantitative analysis}

\paragraph{Hypothesis driven analysis}
\begin{figure*}[tbh]
    \centering
    \includegraphics[width=1\linewidth]{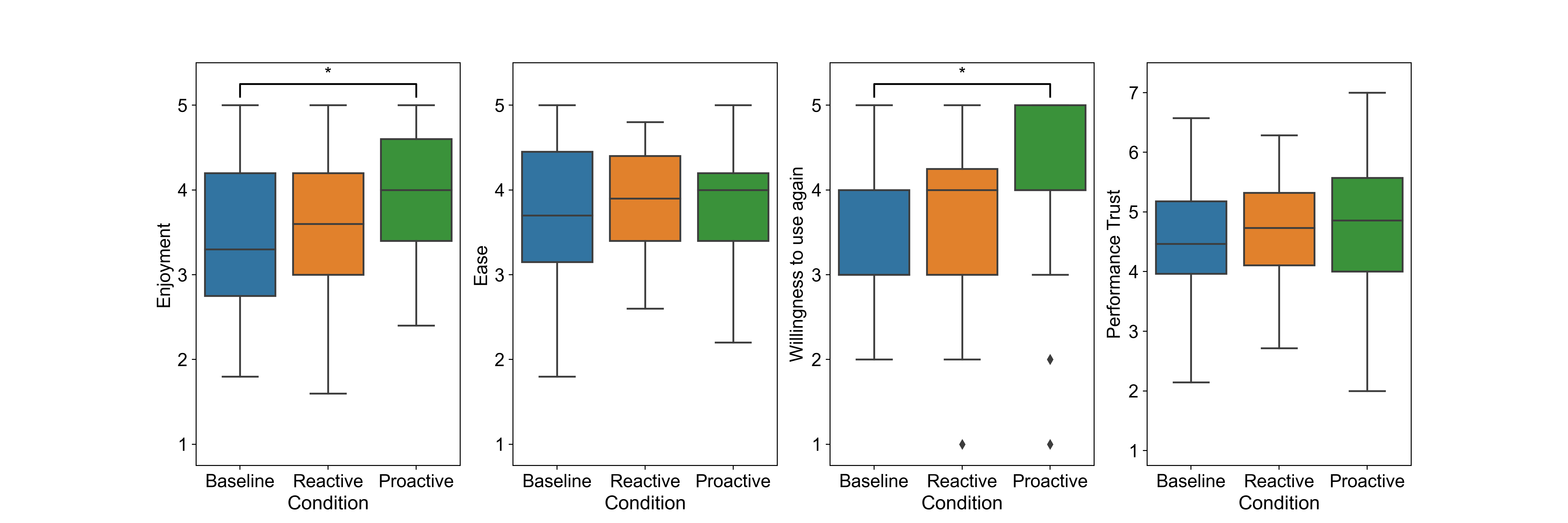}
    \vspace{-2em}
    \caption{The scores for the different scales for the three conditions. Enjoyment, ease of use and willingness to use the robot again are measured on a 5-point Likert scale, while the performance trust uses a 7-point Likert scale.}
    \label{fig:Questionnaire_1}
    \vspace{-2em}
\end{figure*}

We used a MANOVA to evaluate our first questionnaire with respect to our hypotheses. We decided to use the questionnaire ratings collected after the first interaction, since it captures the participant's opinion on the interaction without adding additional bias of them already knowing the questions.
Before running the MANOVA we confirmed that our data is normally distributed and homoscedastic.
Using enjoyment, ease of use, performance trust and willingness to use the robot again as the dependent variables, we compared the different conditions pairwise. For the comparison of the baseline and the reactive condition, no significance was found. The MANOVA for the baseline and the proactive condition shows significant differences ($p <.05$) for all tests.

To gain insights into which dependent variables are different between the two conditions, we use ANOVAs as a post-hoc analysis, with the Benjamini-Hochberg Procedure to account for running multiple tests. The ANOVA showed significance for enjoyment ($F = 5.15$, $p < .05$) and willingness to use the robot again ($F = 1.48$, $p < .05$), but not for ease of use ($F = 0.01$, $p = .95$) and performance trust ($F = 0.17$, $p = .56$). Fig.~\ref{fig:Questionnaire_1} and Tab.~\ref{tab:means} show how users scored the interaction.

\begin{table}[tb]
    \centering
    \caption{The mean (sd) for the scales in the different conditions.}
    \label{tab:means}
    \begin{tabular}{p{0.05\textwidth}p{0.075\textwidth}p{0.075\textwidth}p{0.075\textwidth}p{0.075\textwidth}}
        \toprule
        Condition & Enjoyment & Ease & Willingness to use again & Performance trust \\
        \midrule
        Baseline & 3.34 (0.89) & 3.74 (0.83) & 3.48 (0.93) & 4.59 (0.97) \\
        Reactive & 3.49 (0.90) & 3.88 (0.63) & 3.69 (1.09) & 4.74 (0.88) \\
        Proactive & 3.93 (0.73) & 3.87 (0.71) & 4.00 (1.05) & 4.83 (1.20) \\
        \bottomrule
    \end{tabular}
    \vspace{-2em}
\end{table}

\paragraph{Exploratory analysis}

\begin{figure*}[tbh]
    \centering
    
    \includegraphics[width=0.9\linewidth]{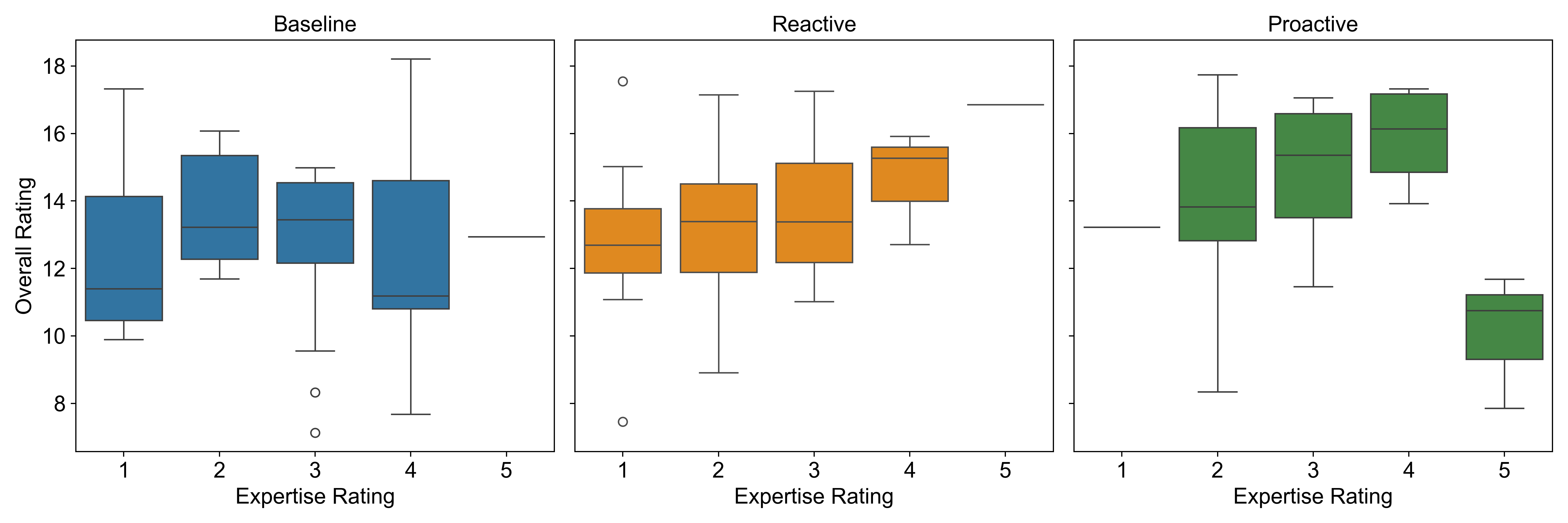}
    \vspace{-1em}
    \caption{The overall scores of the three conditions in regard to the self-reported expertise with social robots (1 = Not familiar at all, 5 = Extremely familiar).}
    \label{fig:experience}
    \vspace{-1em}
    
\end{figure*}

In the beginning of the questionnaires, we asked all participants to rate their experience with social robots on a 5-point Likert scale (Not familiar at all, Slightly familiar, Moderately familiar, Very familiar, Extremely familiar). An analysis of the combined questionnaire scores after normalization with respect to the experience ratings shows that the scores are on a similar level no matter the experience for the baseline (see Fig.~\ref{fig:experience}). However, in the reactive condition, people with more experience give higher overall ratings. In the proactive condition, the overall scores increase first with increasing experience, but there is a steep drop for people who rated themselves \textit{extremely familiar} with social robots. While there are over 36 participants included per condition, there are not many data points for the experience rating of 5.

\begin{figure}[tbh]
    \centering
    \vspace{-2em}
    \includegraphics[width=0.9\linewidth]{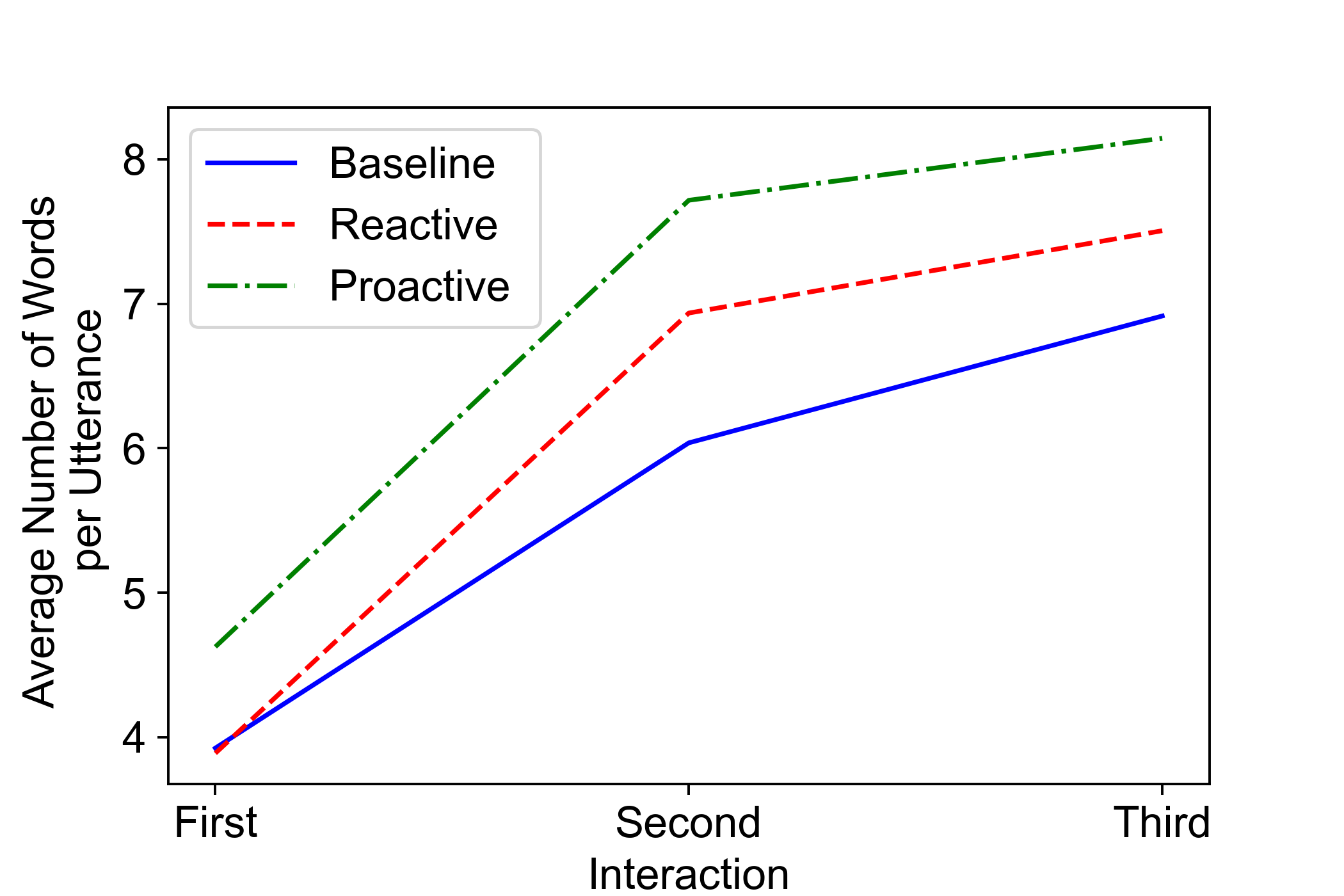}
    \vspace{-0.5em}
    \caption{The average number of user words per utterance per interaction.}
    \label{fig:words}
\end{figure}

In addition to the subjective reports through questionnaires, we collected interaction logs for analysis. The analysis of the number of words per user utterance showed that there was an increase in the number of words used over the three interactions (see Fig.~\ref{fig:words}). In the proactive condition, participants used the most words per utterance for all three interactions (first: M = 4.63 (SD = 3.02), second: M = 7.72 (SD = 6.00), third: 8.15 (SD = 6.57)), which was significantly ($p < .05$) different from the baseline condition. The reactive and baseline conditions started at a similar level (baseline: M = 3.92 (SD = 2.47); reactive: M = 3.89 (SD = 2.72)). However, in the following interactions the reactive condition had a higher average number of words (second: M = 6.94 (SD = 5.71), third: M = 7.51 (SD = 6.321)) than the baseline (second: M = 6.04 (SD = 4.77), third: M = 6.92 (SD = 5.42)).
In contrast to this, the total number of utterances per interaction did not differ significantly between the conditions but increased over the interactions (first: M = 4.91 (SD = 1.54); second: M = 6.88 (SD = 1.94); third: M = 7.37 (SD = 1.59)).

Not only did the participants use more words per utterance in the proactive condition, they also used a greater variety of unique words (Baseline: 548, Reactive: 565, Proactive: 620). One-word utterances were least common in the proactive condition, where they made up 7.69\% of all utterances, compared to 10.64\% in the baseline and 13.29\% in the reactive condition. On a word level, part-of-speech tagging revealed no differences between the conditions.

To see whether the difference in number of words is due to using different dialogue acts, we tagged all utterances using GPT and manual post-processing. We used a total number of 18 tags based on the ISO DA schema \cite{bunt2010towards}, augmented with more task-specific dialogue acts to allow for a more fine-grained analysis (see Tab.~\ref{tab:tags}). In the proactive condition, we observe more informing utterances and information requests instead of direct food requests. In contrast to this, direct food requests make up 20\% of all user utterances in the baseline condition. In the reactive condition, repair actions like (dis)confirmations and clarifications are more common than in the baseline condition. The reactive condition also has the highest number of unclear utterances. A chi-squared test showed significant ($p < .05$) differences between the dialogue act distributions of the conditions.

\begin{table}[tb]
    \centering
    \vspace{-1em}
    \caption{The fraction of user utterances tagged with the corresponding speech acts. The tags \textit{requesting clarification}, \textit{requesting confirmation}, \textit{offering} and \textit{congratulating} were used for less than one percent of the user utterances and are therefore not included in the table. }
    \label{tab:tags}
    \begin{tabular}{lccc}
    \toprule
Tag & Baseline & Reactive & Proactive\\
\midrule
informing & 0.08 & 0.08 & 0.14\\
instructing & 0.03 & 0.01 & 0.02\\
providing confirmation & 0.20 & 0.20 & 0.22\\
disconfirmation & 0.06 & 0.08 & 0.04\\
requesting & 0.00 & 0.01 & 0.00\\
clarifying & 0.03 & 0.05 & 0.03\\
requesting information & 0.13 & 0.13 & 0.16\\
requesting food & 0.20 & 0.18 & 0.15\\
requesting recommendation & 0.07 & 0.05 & 0.05\\
modifying order & 0.04 & 0.04 & 0.03\\
greeting & 0.01 & 0.00 & 0.01\\
apologizing & 0.00 & 0.01 & 0.02\\
thanking & 0.06 & 0.06 & 0.06\\
unclear & 0.09 & 0.12 & 0.09\\
\bottomrule
    \end{tabular}
    \vspace{-2em}
\end{table}

\subsection{Qualitative analysis}
\paragraph{User comments}
After filling in the questionnaires, participants were able to leave additional comments, which was used by 64 participants. 12 participants (baseline: 5, reactive: 5, proactive: 2)
mentioned that the response time was sometimes too long, while 11 (baseline: 1, reactive: 8, proactive: 2) reported speech recognition problems. The combination of the two led to people mentioning that they did not know whether the robot was just processing or had not understood them (``Sometimes it's hard to know whether the robot is taking long to answer or if it didn't catch what I said'' - participant 27, reactive condition).
Although 11 participants reported that they enjoyed the interaction, others found the robot's appearance unsettling and pointed out the robot's eyes and its staring behavior (baseline: 3, reactive: 2, proactive: 4).

Eight people also commented on (not) understanding the robot's behavior (all comments related to understanding are available in the Supplementary Material). Three people in the baseline condition commented that they were uncertain about the understanding, such as this participant who wrote:

\begin{quote}
    ``You're often not sure if it got things right'' - P111, baseline condition.
\end{quote}

The three comments left by participants of the reactive condition in this category were highlighting how confirmations helped with the understanding of the interaction, and that the understanding changed over time. One participant wrote:

\begin{quote}
    ``Interacting with it felt easier over time as I learned about its capabilities.'' - P14, reactive condition
\end{quote}

The same change in understanding over time was reported by a participant in the proactive condition as well.

\begin{quote}
    ``I feel like my experience with the robot changed after using it a couple of times. At the beginning, I was more apprehensive and didn't understand him/her, but after using it again I have more trust in it and enjoyed it more!'' - P109, proactive conditions
\end{quote}

The participants commented on their perceived understanding of the robot's behavior without being told that the study was investigating capability communication.

\paragraph{Speech recognition problems across the conditions}
Since speech recognition problems exist across all conditions, we looked at their distribution to check whether they could be confounding our results. For this we selected 30 interactions (10 of each condition) randomly and analyzed the video data in regard to the number of (severe) misunderstandings that impacted the interaction (baseline: 3, reactive: 7, proactive: 5), non-hearing (baseline: 6, reactive: 4, proactive: 6), latency longer than 7 seconds (baseline: 3, reactive: 0, proactive: 2) and missed sentence beginnings (baseline: 8, reactive: 2, proactive: 0). The missed beginnings happened when the user started talking to early before the robot was listening. Misunderstandings were speech recognition problems, whereas non-hearing could either be the user talking too quietly or early, or the microphone not picking up on the speech.

\paragraph{Interactions with the lowest ratings}
To get insights into the problems that affected the interactions in the different conditions, we selected the five interactions with the lowest ratings from each condition and looked at the participants' comments, as well as the interaction recordings. The inspection of the interaction recordings showed that the robot not picking up on the user's speech at all was an issue in all three conditions (baseline: 13, reactive: 9, proactive: 12) together with general speech recognition problems (baseline: 5, reactive: 3, proactive: 5). However, while five (reactive) and four (proactive) interactions ended successfully with the user getting the complete order they wanted in the capability communication conditions, only two interactions from the baseline condition achieved this task success.

The comments of the five lowest-rated baseline interactions mentioned a lack of emotions from the robot, the need for exaggerated speech to be heard at all, and not knowing why certain things were happening. The proactive conditions' comments criticized the rigid structure of the conversation, and the speech recognition, while also mentioning frustration and that confirmations helped whereas no confirmations made it more difficult. The proactive condition had a rather positive comment, describing the interaction as ``somehow funny'' (P1, proactive condition), while the negative comments also mentioned the overarticulation. Additionally, one participant emphasized their unhappiness with the robot's appearance: ``It was weird. The way the robot moved and looked was scary'' - P63, proactive condition.

Although people in all conditions reported that the interaction suffered from speech recognition problems, the interaction recordings showed that the participants reacted differently when the robot did not hear. Participants of the proactive condition repeated their utterance, often visibly frustrated. In contrast to this, participants from the baseline condition displayed confusion about why the robot was not reacting (correctly) and just waited for a reaction from the robot's side instead of attempting to repair directly.

\section{Discussion}

\subsection{Subjective experience}
The questionnaire results showed that people preferred the interaction with the proactive robot over the baseline, in regard to enjoyment, and that most of them were willing to use it again, supporting our second hypothesis. However, we did not find significant differences between the baseline and reactive condition (H1) and the reactive and proactive condition (H3). A factor that seems to have influenced the user's perception of the reactive condition was speech recognition problems. 12\% of all utterances in the condition were tagged as unclear, and eight participants reported speech recognition problems in the final comments. This problem with the interaction is also reflected in the dialogue acts, as there are more disconfirmations and clarifications in the reactive condition, indicating that the robot was not correctly responding to what the person had said. In the randomly sampled 10\% of interactions, which were analyzed regarding the speech recognition problems, we found more than twice the amount of misunderstandings for the reactive condition than for the baseline. While missed beginnings and non-hearing are influenced by the user's timing, misunderstandings are because of the failures of the robot. It is generally difficult to design a study with failures, if the failures should happen naturally. While we do expect failures to happen in an autonomous interaction, we cannot fully control when and where they are going to happen.

Although there was a significant overall difference between the baseline and the proactive condition, a more fine-grained analysis revealed that only enjoyment and willingness to use the robot again were significantly different. In the past, some studies found that transparency positively influenced ease of use, while others did not find an effect \cite{Bhaskara_Skinner_Loft_2020}. Our study is more in line with the latter, with no significant difference in perceived ease of use ratings. However, the open-ended questions indicated that people felt like using the robot became easier over time in the reactive and proactive conditions, while nobody mentioned it for the baseline condition. The lack of difference for ease of use could also originate from the users having difficulties estimating whether an interaction is easy or not. Especially people who have not interacted with a robot before, do not have a baseline value for how difficult an interaction with a robot is.

The strategies are suitable for different expertise levels, as indicated by the overall interaction ratings. Although expertise does not appear to matter in the baseline condition, participants with higher expertise scored the reactive condition higher than those with lower expertise. A possible explanation for this is that experienced users use the robot's explanations and information to make the conversation smoother and more enjoyable for them. Inexperienced users show a greater variety in scores, which could be attributed to them not knowing how to use the additional information due to being unfamiliar with how to interact with robots in the first place. Additionally, the reactive condition did not provide users with unnecessary information, if they had smoother interactions due to their expertise. This information that the participants do not need and want is also a possible explanation for the sudden drop in the proactive condition, where the participant got the information every interaction regardless of how it was going.

The comments about the understanding of the robot's behavior left at the end of the study indicated that the understanding of what they can do during the interaction did depend on a mix of the robot's behavior and its communication. While participants in the baseline condition only mentioned the problem of not knowing whether the robot understood them, participants from the other conditions also commented that they were able to learn about the robot's behavior over time.

\subsection{User behavior}
The user behavior differed between the conditions, with the number of words per utterance used being one of the differences (see Fig.~\ref{fig:words}). 
In the proactive condition, the users did already use more words per utterance during the first interaction, which can be explained by the robot starting the interaction with explanations about its capabilities. Those explanations allow users to already make some first-grounded assumptions before learning about the interaction through interacting. The behavior change is also reflected in the increased number of words per utterance for all conditions over the interactions. The increase is faster in the conditions where the user got additional information about the robot's capabilities, indicating that the capability communication achieved its purpose of helping the user understand how they can interact with the robot. The fact that all conditions showed an increase in the number of words per utterance over the interactions, indicates that people generally preferred an interaction style with more words, once they were more familiar with the robot.

Since the interaction with the ARI robot was speech-based, users could use different interaction styles, which became apparent not only in the length of the utterances but also in the different dialogue acts used. The users could, for example, either adopt a command-like style or interact more conversationally, similar to how they would with a human working in a restaurant. The robot was able to deal with such conversational interactions. Using more words in a sentence showed the confidence of the user that the robot would be able to handle this more conversational interaction style.

Having fewer one-word utterances in the proactive condition is also reflected in the dialogue acts used (see Table~\ref{tab:tags}). Instead of \textit{requesting food} directly (``Pizza'' - P51, baseline condition), the participants of the proactive condition used more statements that were \textit{informing} or \textit{requesting information} (``one of my friends is gluten-free and vegan do you have anything for that'' - P2, proactive condition). Together with using more unique words, this suggests that users in the proactive condition do not limit their way of talking with the robot to commands and short utterances, but go for a more conversational interaction style instead.

The difference in user behavior is visible in the lowest-rated interactions as well. Participants in the baseline condition seemed confused about the robot's behavior when something went wrong, while those in the proactive condition showed signs of frustration. The confusion in the baseline condition led to people being more hesitant to initiate repair directly, as they were unsure what had gone wrong. This highlights that it is important for people to get, at least to a certain degree, insights into the cause of robot behaviors to understand them, which is also well-known from literature \cite{Schött_Amin_Butz_2023}. 
However, the reaction of the participants in the proactive condition, whose behavior indicated an understanding of the problem, showed that the nature of the problem itself can also lead to frustration. The following example (P105, reactive condition) shows such an instance of a participant having to clarify a speech recognition problem, which was not relevant:
\begin{quote}
\textbf{User:} could you make it quick please find 500\\
\textbf{Robot:} Of course, we will ensure your order is prepared quickly. However, I didn't understand the "find 500" part. Could you please clarify?\\
\textbf{User:} I just asked if you could make it quick I am quite hungry
\end{quote}
Therefore, the reactive condition can decrease the illusion that things are going well, and by that, also lead to decreased satisfaction \cite{fischer2018transparent}.

Although we tried to reduce speech recognition problems by running the study in a quiet environment and by using both intent classification and automatic transcript correction when the intent recognition failed, around 10\% of all utterances were not recognized. Speech recognition problems where the robot was not picking up any sounds are not included in this statistic, since the robot could not create transcripts of something it did not perceive. In the reactive condition, the robot communicated only about its capabilities when problems were detected, which means that there was no capability communication if the problem was that the robot did not hear anything. In contrast to that, the robot always communicated its capabilities in the proactive condition, which could be a reason for the big difference between the two, since the users did not rely on the detection of a problem by the robot to receive this information. We looked at a random subset of the interactions and saw that misunderstandings and non-hearing were present in all conditions. The high number of missed sentence beginnings in the baseline condition indicates that the participants were misjudging the robot and when it was listening to them.

\subsection{Limitations}
The speech recognition problems and the occasional latency are limiting factors in our study, affecting all conditions. The reactive condition was only triggered if the robot was able to detect the problem, meaning that it did not account for times when the robot did not pick up the participant's speech. Using an additional microphone and not just ARI's, might offer options to also account for those cases. While we looked at 33 randomly selected interactions to look at the speech recognition problems, it might be beneficial to also analyze those for the rest of the data. Wile we tried to minimize the effect of running our study in the lab by using a kitchen lab and observing the interaction from an extra observation room, it could still have an impact on the interaction.

\section{Conclusion}
Our study on capability communication in human-robot interaction shows that users prefer a robot that proactively communicates its capabilities over one that does not communicate them. This is reflected not only in higher overall questionnaire ratings but also in the way humans interact with the robot by adopting a more conversational interaction style.
However, our results also indicate that the user's experience with social robots also plays a role in how much they enjoy using systems with different capability communication strategies. For speech-based HRI, it makes, therefore, sense to not just pay attention to the content of the interaction itself, but also adapt the robot's capability communication to the user.

For future work, it would be interesting to see how the findings translate to long-term interactions and whether it makes sense to adapt the capability of communication over multiple interactions.




\bibliographystyle{IEEEtran}
\bibliography{bibliography}

\newpage

\section*{Appendix}

\subsection{Tools and Materials}
\begin{itemize}
    \item Robot: ARI (1.6 m tall humanoid robot of PAL Robotics)
    \item Speech recognition: Google Dialogflow
    \item Intent Detection: Google Dialogflow (agent included in supplementary material)
    \item Transcript correction and NLG: GPT-4 (prompts included in supplementary material)
    \begin{itemize}
        \item confidence value of 0.45 for assuming speech recognition problems
    \end{itemize}
    \item Questionnaires:
    \begin{itemize}
        \item Enjoyment: \href{https://link.springer.com/content/pdf/10.1007/s12369-010-0068-5.pdf}{PENJ subscale}
        \item Ease of Use: \href{https://link.springer.com/content/pdf/10.1007/s12369-010-0068-5.pdf}{PEOU subscale}
        \item Performance trust: \href{https://research.clps.brown.edu/SocCogSci/Measures/CurrentVersion_MDMT.pdf}{MDMT}
    \end{itemize}
\end{itemize}

\subsection{User Comments}

\begin{table}[hb]
    \caption{Comments focusing on the understanding of the robot's behavior and interaction.}
    \label{tab:comments}
    \begin{tabular}{p{0.075\textwidth}cp{0.75\textwidth}}
        \toprule
        \textbf{Participant Nr.} &\textbf{Condition} & \textbf{Comment} \\
        \midrule
        98 & Baseline & Overall good service, but unreliable to know which language can be used. \\
        111 & Baseline & You're often not sure if it got things right \\
        114 & Baseline & I dont know what happened as it didnt get my orders correct and then I couldn't tell if it actually made the special order vegan or not, so I think I it was more frustrating by the end. \\
        \midrule
        14 & Reactive & Interacting with it felt easier over time as I learned about its capabilities. \\
        75 & Reactive & I would need more time to see the result my order to see if it understood I maybe need a manual in beginning I find it interesting and look forward to try reality \\
        86 & Reactive & Lack of confirmation from the robot on some of my answers made it hard to focus on what happened next. The confirmation made it easier to notice what was happening faster. \\
        \midrule
        65 & Proactive & If I was in a restaurant, and I knew that the robot is supposed to work correctly 100 percent of the times, I would know when the robot is working or not. Because I am not sure what is being tested, I am not sure if the robot not understanding me is part of its test or if I was doing something wrong. \\
        109 & Proactive & I feel like my experience with the robot changed after using it a couple of times. At the beginning I was more apprehensive and didnt understand him/her, but after using it again I have more trust in it and enjoyed it more! \\
        \bottomrule
    \end{tabular}
\end{table}

\end{document}